\documentclass{article}
\usepackage[utf8]{inputenc}
\usepackage{amsmath}
\usepackage{amssymb}

\title{An Algorithm for Approximating Continuous Functions on Compact Subsets with a Neural Network with one Hidden Layer}
\author{Elliott Zaresky-Williams}
\date{February 7th 2019}
\usepackage{biblatex}
\usepackage{graphicx}

\begin{document}

\maketitle
\begin{abstract}
    George Cybenko's landmark 1989 paper showed that there exists a feedforward neural network, with exactly one hidden layer (and a finite number of neurons), that can arbitrarily approximate a given continuous function $f$ on the unit hypercube. The paper did not address how to \textit{find} the weight/parameters of such a network, or if finding them would be computationally feasible. This paper outlines an algorithm for a neural network with exactly one hidden layer to reconstruct any continuous scalar or vector valued continuous function. 
    
\end{abstract}
\section{Introduction}

Cybenko proved for every $f\in C([0,1]^n)$, we have that functions of the form $P(x)=\sum_{j=1}^N \alpha_j\sigma(y_j^{T}x+\theta_j)$ approximate $f$ arbitrarily well with $||P-f||_{\infty}<\epsilon$. (1). Such functions are known to be implementable on a feedforward neural network with one hidden layer, with the assumption the activation function(s) are bounded and non-constant. (2) 

This result was a significant theoretical milestone. In principle, a neural network could learn a highly non-linear decision boundary for data classification, making them a preferred alternative to SVMs, which are limited by a specific choice of kernel function. In addition, any continuous function that models a stochastic process (such as Brownian Motion/Wiener Process) could be modeled by some neural network. 

In unsupervised learning, neural networks can learn arbitrary probability distributions $\mathbf{P}(X)$ of a given data set (as is done with GANs). 

Cybenko proved the \textit{existence} of a neural network with one hidden layer which could approximate a given continuous function of interest. How we actually find  the weights/parameters of the network was not addressed. Gradient descent is a common method to find the weights, but gradient descent does not touch on how well the network will learn an arbitrary function, how many training iterations it will take, or how many artificial neurons in the hidden layer to use. Additionally, the specific choice of activation function plays a key role in approximating a given continuous function. All of these questions will be addressed in this paper through a single theorem, which allows us to train a neural network without using an iterative method. 

The next section will state the main theorem (and its proof) outright. Then, the hypotheses of the theorem will be carefully examined, followed by a practical explanation and applications to various domains in machine learning.
Since the theorem provides an algorithmic way we can approximate a continuous function, it will also be referred to as the "Universal Function Algorithm" (UFA for short). 

\section{Main Theorem and Discussion}

This paper will use functions of a slightly different form to represent neural networks with one hidden layer.

Let $x$ denote the input into the neural network. Let $\delta$ denote the weight into the hidden node $\alpha$. Let $g$ denote the activation function into the hidden layer. We have $g(x\delta)=\alpha$. 

For the output layer, denote the weight into the output node $y$ as $\theta$, and let $\sigma$ denote the activation into the output node. Mathematically, we have $y=\sigma(\alpha \theta)$. To reconstruct any ordered pair $(x_0,f(x_0))$ with a neural network, we only need $x_0$ as the input node, one hidden node, and one output node $f(x_0)$.

The goal will be to show that for any $f\in C([a,b])$, we have $$f(x_0)=y=\sigma(g(x_0\delta)\theta)$$

for some $\delta,\theta \in \mathbb{R}$ and an arbitrary $x_0\in [a,b]$. Since we will be able to do this for every $x_0\in [a,b]$, we can can reconstruct $f$ at every point. The theorem will now be presented. 

\subsection{Universal Function Approximation Theorem} Let $f,g\in C([a,b])$. Let  $\sigma\in C^1([a,b])$ with $g, \sigma'\neq 0$ $\forall x \in [a,b]$ and let $f([a,b])\subseteq \sigma([a,b])$. Then $\forall x\in [a,b]$,  $\exists \theta\in \mathbb{R}$ $\forall \delta \in \mathbb{R}$, where $$f(x)=y=\sigma(g(x\delta)\theta)$$. 

\subsubsection{Proof}

By assumption, we known that $\sigma$ is invertible since it is $C^1([a,b])$ with $\sigma'\neq 0$  $\forall x \in [a,b]$. Inverting the function yields:

$\sigma^{-1}(f(x))=g(x\delta)\theta$. Now, we just set

$$\theta=\frac{\sigma^{-1}(f(x))}{g(x\delta)}$$

It remains to verify that this is indeed the correct parameter:

$$f(x)=\sigma(g(x\delta)(\frac{\sigma^{-1}(f(x))}{g(x\delta)}))$$

$$f(x)=\sigma(\sigma^{-1}(f(x)))$$

We have found a $\theta$ for each $x$. Since the above holds $\forall x\in [a,b]$, we are done.

\subsubsection{Discussion of Theorem and Assumptions}

The assumption that $f([a,b])\subseteq \sigma([a,b])$ was vital. As an example, suppose we wanted the network to learn the output $y=2$ using only a sigmoid  function $\sigma(x)=\frac{1}{1+exp(-x)}$ with range $(0,1)$. The image of $f$ is not contained within $\sigma$, and so there is no weight that will lead the network to learn the output value of $2$. 

In addition, we see that if $\sigma'$ vanishes for some $x$ in $[a,b]$ then $\sigma$ is no longer invertible at $x$, and we lose the guarantee of an optimal weight. Similarly, if $g$ vanishes for some $x\in [a,b]$, then $\theta$ is undefined at $x$. 

It may be tempting to extend the results to $f\in L^1([a,b])$ via the method above. Hornik did show that a multilayer feedforward network could actually learn any \textit{measurable} function (2). Even more generally, neural networks have been shown to be able to approximate some more general maps between compact groups (3). 

Unfortunately, there isn't an obvious way to extend the result to even $L^p([a,b]^n)$ functions because if 
$f(x)=\sigma(g(x\delta)\theta)$, $f$ must also be continuous since $\sigma$ is itself continuous. 

Finally, it is possible to treat the parameters $\delta,\theta$ as functions of $x$. In the main theorem, when writing $f(x)=\sigma(g(x\delta)\theta)$ as above, an $x\in [a,b]$ is fixed. The weight $\theta$ is then computed for that \textit{specific} $x$ for an arbitrary fixed $\delta\in \mathbb{R}$. The "pointwise" version of the main theorem has an alternative formulation. 

\subsubsection{Alternative Statement of Theorem}
Let $f,g\in C([a,b])$, $\sigma\in C^1([a,b])$. Let $g,\sigma'\neq 0 \hspace{2mm} \forall x\in [a,b]$. Then, $\exists \delta, \Theta\in C([a,b])$ where $f(x)=\sigma(g(x\delta(x))\Theta(x))$ is satisfied $\forall x\in [a,b]$. 

The proof of this theorem follows exactly the same logic as the original version. Here, we compute the weight function $\Theta(x)$ as

$$\Theta(x)=\frac{\sigma^{-1}(f(x))}{g(x\delta(x))}$$ where $\delta(x)$ can be an arbitrary $C([a,b])$ function. We see here that both versions of the theorem are essentially the same. 

For the rest of the paper, when referring to the UFA/main theorem, the "pointwise" version (as in the original statement) will be used.

\subsection{Extension to scalar valued functions on $[a,b]^n$}

The theorem extends to $f\in C([a,b]^n)$. Here, the neural network takes a vector $x\in \mathbb{R}^n$ as input. The hidden layer still only has one node per input $x$, and the output layer has one node.  

A function $f\in C([a,b]^n)$ can be represented as
$$f(x)=\sigma(g(\sum_{k=1}^n x_k\delta_k)\theta)$$

The conditions/assumptions are the same as the main theorem, with the difference being we compute a $\theta$ for a fixed vector $x\in \mathbb{R}^n$. This time, the optimal weight is $$\theta=\frac{\sigma^{-1}(f(x))}{g(\sum_{k=1}^n x_k\delta_k)}$$

Following the algebraic steps as in the earlier proof, we see this is indeed the correct value for $\theta$. 

\subsection{Extension to vector valued functions}

Suppose we want to approximate a continuous function $f\in C([a,b]^n,[a,b]^m)$. In the language of neural networks, this network takes $n$ inputs and produces $m$ outputs, or $f(x_1,x_2,...,x_n)=(y_1,y_2,...,y_m)$. 

$\forall j$ with  $1\leq j \leq m$, we have $$y_j=\sigma_j(g(\sum_{k=1}^n x_k\delta_k)\theta_j)$$ As an example, if we had a function $f:[a,b]^{10}\rightarrow [a,b]^2$, the neural network representing that function would be a 2-tuple:
$$Y=(y_1,y_2)$$

$$Y=(\sigma_1(g(\sum_{k=1}^{10} x_k\delta_k)\theta_1), \hspace{2mm} \sigma_2(g(\sum_{k=1}^{10} x_k\delta_k)\theta_2))$$

For each $y_j$ in our output vector, we treat $\sigma_j$ as a scalar valued activation function, with the same conditions as before. Here, the $\sigma_j$'s can all be distinct, as long as each one satisfies the conditions of the main theorem. As expected, the weights $\theta_j$ are given by:

$$\theta_j=\frac{\sigma^{-1}_j(y_j)}{g(\sum_{k=1}^n x_k\delta_k)}$$. 

The vector-valued version of the main theorem is nuanced. The input from the hidden layer to each output node $y_j$ is scalar valued, but the overall output layer can be interpreted as a vector in $\mathbb{R}^m$. Essentially, we are applying the multidimensional scalar valued version of the theorem $m$ times for an output vector $Y\in \mathbb{R}^m$. 

This formulation is particularly advantageous because $\forall j$, $\sigma_j$ is a function of a single real variable. In general, it is easier to find the inverse $\sigma^{-1}$ of a single variable function than a multivariate function. 

\subsection{Approximation vs. perfect reconstruction}

In order to \textit{perfectly} reconstruct $f\in C([a,b]^n)$, we would need an uncountable number of input and output nodes. Intuitively, we have a function we want to reconstruct $f$, but we start with an activation function that needs to be appropriately scaled to match the $f$. 

However, if we know $f$ at $p$ distinct (i.e. discrete) points then we can apply the UFA at each known point (or vector) $p$ for $1\leq i\leq p$ to perfectly reconstruct $f$ at the known points. In any numerical application, we only can sample at a finite number of points, so for practical purposes, we are able to reconstruct $f$ completely in the discrete setting. We can make this comment more precise. 

Let $f\in C([a,b]^n,[a,b]^m)$, with $n,m\geq 1$, and let $f$ be sampled at $p$ distinct points. The neural network needed to perfectly reconstruct $f$ has $np$ input nodes ($n$-dimensional input vector for each of the $p$ points in the domain), $p$ hidden nodes (a hidden node for each of the input vectors), and $mp$ output nodes ($m$-dimensional output vector for each point $p$). In other words, an $n$-dimensional input vector feeds into one hidden node, and the hidden node has $m$ weights associated to the $m$-dimensional output vector. This architecture repeats for every point/vector $p$.

\subsection{Strengths and Weaknesses}

 The performance of a neural network on supervised learning tasks is measured by a loss function, which measures how "far" the network output is to the desired/target output. Let $\hat y$ be target output and $y$ be the output the neural network generates, and $L(\hat y,y)$ the loss function ($\hat y, y$ are vectors in general). The loss function can take many forms (Least-squares, cross-entropy etc.), but for any loss function, $L(\hat y, y)=0$ iff $\hat y=y$. Since a shallow network can always attain $\hat y$ (under the assumptions given above), it immediately follows the loss function $L(\hat y, y)$ is not only minimized, but it is guaranteed to be $0$. For universal function approximation, this algorithm is superior to gradient descent for several reasons. 
 
First, UFA (assuming conditions are met) guarantees we achieve zero loss. There is no guarantee with gradient descent. Depending on activation function chosen, the target functions we are interested in, and the number of nodes chosen, we may or may not get low loss. Second, UFA also is not iterative, it can train in a single step and the optimal architecture for the neural network to implement the UFA is known. Finally, gradient descent has an additional parameter to initialize before training: the learning rate/step size, which affects the speed at which the global minimum is attained (if it is attained at all). A priori, we do not know beforehand what the learning rate should be set to. 

Unfortunately, UFA is not practical outside function approximation. There is no obvious way to extend this to data classification effectively. It's likely that the UFA will lead to overfit models since it could perfectly learn to classify the training data, but in doing so, capture all the noise of the training set as well, ultimately leading to poor performance on the test set.

\section{Conclusion}

A constructive algorithm for finding the weights in a neural network needed to reconstruct an arbitrary continuous function was shown. While direct practical applications are not immediately obvious, we see that shallow networks can model arbitrarily complex processes, as long as the processes can be fully modeled by some continuous function on a compact subset of $\mathbb{R}^n$, as first shown by Cybenko in 1989.

\end{document}